\documentclass{article} % For LaTeX2e
\pdfoutput=1
\usepackage{nips12submit_e,times}
\usepackage[algo2e,ruled,noline]{algorithm2e}
\usepackage{graphicx}
\usepackage{amssymb}
\usepackage{amsmath}
\usepackage{amsthm}
\usepackage{multirow,colordvi} 
\newcommand{\argmax}{\operatornamewithlimits{argmax}}
\usepackage{lscape,xfrac}

\DeclareMathOperator{\mP}{\mathcal{P}}

\DeclareMathOperator{\mD}{\mathcal{D}}

\DeclareMathOperator{\tp}{\tilde{p}}
\DeclareMathOperator{\tq}{\tilde{q}}

\DeclareMathOperator{\tbq}{\mathbf{\tilde{q}}}
\DeclareMathOperator{\tbp}{\mathbf{\tilde{p}}}
\DeclareMathOperator{\bp}{\mathbf{p}}
\DeclareMathOperator{\bq}{\mathbf{q}}

\usepackage{dsfont}

\makeatletter
\renewcommand\paragraph{\@startsection{paragraph}{4}{\z@}%
	{-3.25ex\@plus -1ex \@minus -.2ex}%
	{1.5ex \@plus .2ex}%
	{\normalfont\normalsize\bfseries}}
\makeatother
\usepackage{subfigure}

\title{Efficient Bayes-Adaptive Reinforcement Learning using Sample-Based Search}

\author{
Arthur Guez\\
\texttt{\footnotesize{aguez@gatsby.ucl.ac.uk}} \\
\And
David Silver\\
\texttt{\footnotesize{d.silver@cs.ucl.ac.uk}} \\
\And
Peter Dayan\\
\texttt{\footnotesize{dayan@gatsby.ucl.ac.uk}} \\
}

\nipsfinalcopy % Uncomment for camera-ready version

\begin{document}
\maketitle

%%%%%%%%%%%%%%%%%%%%%%%%%%%%%%%%%%%%%%%%%%%
%
%   ABSTRACT
%
%
%%%%%%%%%%%%%%%%%%%%%%%%%%%%%%%%%%%%%%%%%%%
\vspace{-0.1in}
\begin{abstract}
Bayesian model-based reinforcement learning is a formally elegant
approach to learning optimal behaviour under model uncertainty,
trading off exploration and exploitation in an ideal
way. Unfortunately, finding the resulting Bayes-optimal policies is
notoriously taxing, since the search space becomes enormous. In this
paper we introduce a tractable, sample-based method for approximate
Bayes-optimal planning which exploits Monte-Carlo tree search. Our
approach outperformed prior Bayesian model-based RL algorithms by a
significant margin on several well-known benchmark problems -- because
it avoids expensive applications of Bayes rule within the search tree
by lazily sampling models from the current beliefs. We illustrate the
advantages of our approach by showing it working in an infinite state
space domain which is qualitatively out of reach of almost all previous
work in Bayesian exploration.
\end{abstract}

%%%%%%%%%%%%%%%%%%%%%%%%%%%%%%%%%%%%%%%%%%%
%
%   INTRODUCTION
%
%
%%%%%%%%%%%%%%%%%%%%%%%%%%%%%%%%%%%%%%%%%%%
\section{Introduction}
\vspace{-0.04in}

A key objective in the theory of Markov Decision Processes (MDPs) is
to maximize the expected sum of discounted rewards when the dynamics
of the MDP are (perhaps partially) unknown. The discount factor
pressures the agent to favor short-term rewards, but potentially
costly exploration may identify better rewards in the long-term. This
conflict leads to the well-known exploration-exploitation trade-off.
One way to solve this dilemma 
\cite{Bellman:1959:aaa, Feldbaum:1960:aaa} is to augment the regular state of
the agent with the information it has acquired about the dynamics. One formulation of this idea is the augmented Bayes-Adaptive MDP (BAMDP)
\cite{Martin:1967:aaa,Duff:2002:aaa}, in which the extra information
is the posterior belief distribution over the dynamics, given the data so far
observed. The agent starts in the belief state corresponding to its
prior and, by executing the greedy policy in the BAMDP whilst updating its posterior, acts optimally (with respect to its beliefs) in the original MDP. In this framework, rich prior knowledge about
statistics of the environment can be naturally incorporated into the
planning process, potentially leading to more efficient exploration
and exploitation of the uncertain world.
 
Unfortunately, exact Bayesian reinforcement learning is
computationally intractable. Various algorithms have been devised to
approximate optimal learning, but often at rather large cost. Here, we
present a tractable approach that exploits and extends recent advances in
Monte-Carlo tree search (MCTS) \cite{Kocsis:2006:aaa, Silver:2010:aab}, but
avoiding problems associated with applying MCTS  directly to the BAMDP. 

At each iteration in our algorithm, a single MDP is sampled from the
agent's current beliefs. This MDP is used to simulate a single episode
whose outcome is used to update the value of each node of the search
tree traversed during the simulation.  By integrating over many
simulations, and therefore many sample MDPs, the optimal value of each
future sequence is obtained with respect to the agent's beliefs. We
prove that this process converges to the Bayes-optimal policy, given
infinite samples. To increase computational efficiency, we introduce a
further innovation: a lazy sampling scheme that considerably reduces
the cost of sampling.

We applied our algorithm to a representative sample of benchmark
problems and competitive algorithms from the literature. It
consistently and significantly outperformed existing Bayesian RL
methods, and also recent non-Bayesian approaches, thus achieving 
state-of-the-art performance.\footnote{Note: an extended version of this paper 
is available, see reference \cite{Guez:2013:aaa}.}

Our algorithm is more efficient than previous sparse sampling methods
for Bayes-adaptive planning \cite{Wang:2005:aaa, Castro:2007:aaa,
Asmuth:2011:aaa}, partly because it does not update the
posterior belief state during the course of each simulation. It thus
avoids repeated applications of Bayes rule, which is expensive for all
but the simplest priors over the MDP. Consequently, our algorithm is
particularly well suited to support planning in domains with richly
structured prior knowledge --- a critical requirement for
applications of Bayesian reinforcement learning to large problems. We
illustrate this benefit by showing that our algorithm can tackle a
domain with an infinite number of states and a structured prior over
the dynamics, a challenging --- if not intractable --- task for
existing approaches.

%%%%%%%%%%%%%%%%%%%%%%%%%%%%%%%%%%%%%%%%%%%
%
%   BAYESIAN RL Formalism
%
%
%%%%%%%%%%%%%%%%%%%%%%%%%%%%%%%%%%%%%%%%%%%
\section{Bayesian RL}
\label{sec:brl}

We describe the generic Bayesian formulation of optimal decision-making in an
unknown MDP, following \cite{Martin:1967:aaa} and \cite{Duff:2002:aaa}. An MDP
is described as a 5-tuple $M = \langle S, A,\mathcal{P},\mathcal{R},\gamma
\rangle$, where $S$ is the set of states, $A$ is the set of actions,
$\mathcal{P} : S \times A \times S \rightarrow \mathbb{R}$ is the state
transition probability kernel, $\mathcal{R} : S \times A \rightarrow
\mathbb{R}$ is a bounded reward function, and $\gamma$ is the discount 
factor~\cite{Szepesvari:2010:aaa}.  When all the components of the MDP tuple are
known, standard MDP planning algorithms can be used to estimate the optimal value function and
policy off-line. In general, the dynamics are unknown, and we assume that $\mathcal{P}$ is a
latent variable distributed according to a distribution $P(\mathcal{P})$.
After observing a history of actions and states $h_t=s_1a_1s_2 a_2
\dots a_{t-1} s_t$ from the MDP, the posterior belief on 
$\mathcal{P}$ is updated using Bayes' rule $P(\mathcal{P}| h_{t}) \propto P(h_{t}
| \mathcal{P}) P(\mathcal{P})$. The uncertainty about the dynamics of the
model can be transformed into uncertainty about the current state inside an
augmented state space $S^+ = S \times \mathcal{H}$, where $S$ is the state space in
the original problem and $\mathcal{H}$ is the set of possible histories. 
%Make a note here that histories are not the most compact sufficient statistics
The dynamics associated with this augmented state space are described by 
\begin{equation}
\mathcal{P}^+(\langle s,h \rangle, a, \langle s',h' \rangle) = 
	\mathds{1}[h' = has']
	\int_{\mathcal{P}} \mathcal{P}(s,a,s') P(\mathcal{P} | h) \, \text{d} \mathcal{P}, \;\;\;\;
\mathcal{R}^+( \langle s,h \rangle ,a) = R(s,a)
\label{eq:BAMDP}
\end{equation}
Together, the 5-tuple $M^+ = \langle S^+, A,\mathcal{P}^+,\mathcal{R}^+,\gamma
\rangle$ forms the Bayes-Adaptive MDP (BAMDP) for the MDP problem $M$. Since
the dynamics of the BAMDP are known, it can in principle be solved to obtain
the optimal value function associated with each action:
%using Bellman'sequation  to obtain the optimal value function: 
\begin{equation}
%\begin{split}
	Q^*(\langle s_t,h_t \rangle, a) = \max_\pi \mathbb{E}_\pi \left[ \sum_{t'=t}^{\infty} \gamma^{t'-t} r_{t'} | a_t = a\right] 
%\max_a \mathcal{R}^+(\langle s,h \rangle,a) \\
%&+ \gamma \int_{\langle s',h' \rangle} \mathcal{P}^+( \langle s,h \rangle, a, \langle s',h' \rangle) V^*(\langle s',h' \rangle)  
%\end{split}
\label{eq:optQ}
\end{equation} 
from which the optimal action for each state can be
readily derived.~\footnote{The redundancy in the state-history tuple notation --- $s_t$ is the suffix of $h_t$ ---
is only present to ensure clarity of exposition.}
Optimal actions in the BAMDP are executed greedily in the real MDP $M$
and constitute the best course of action for a Bayesian agent with
respect to its prior belief over $\mathcal{P}$. It is obvious that the
expected performance of the BAMDP policy in the MDP $M$ is bounded
above by that of the optimal policy obtained with a
fully-observable model, with equality occurring, for example, in the
degenerate case in which the prior only has support on the true model.

%%%%%%%%%%%%%%%%%%%%%%%%%%%%%%%%%%%%%%%%%%%
%
%  BAMCP
%
%
%%%%%%%%%%%%%%%%%%%%%%%%%%%%%%%%%%%%%%%%%%%
\section{The BAMCP algorithm}
\label{sec:bmcp}

\subsection{Algorithm Description}

The goal of a BAMDP planning method is to find, for each decision
point $\langle s, h \rangle$ encountered, the action $a$ that
maximizes Equation~\ref{eq:optQ}.  Our algorithm, Bayes-adaptive Monte-Carlo Planning (BAMCP),
does this by performing a forward-search in the space of possible
future histories of the BAMDP using a tailored Monte-Carlo tree search.  

We employ the UCT algorithm \cite{Kocsis:2006:aaa} to allocate search
effort to promising branches of the state-action tree, and use
sample-based rollouts to provide value estimates at each node. For clarity, let
us denote by Bayes-Adaptive UCT (BA-UCT) the algorithm that applies vanilla UCT
to the BAMDP (i.e., the particular MDP with dynamics described in
Equation~\ref{eq:BAMDP}). Sample-based search in the BAMDP using BA-UCT
requires the generation of samples from $\mathcal{P}^+$ at every single node. 
This operation requires integration over all possible transition models, or
at least a sample of a transition model $\mathcal{P}$ --- an expensive
procedure for all but the simplest generative models
$P(\mathcal{P})$. We avoid this cost by only sampling a single
transition model $\mathcal{P}^i$ from the posterior at the root of the
search tree at the start of each simulation $i$, and using
$\mathcal{P}^i$ to generate all the necessary samples during this
simulation. Sample-based tree search then acts as a filter, ensuring
that the correct distribution of state successors is obtained at each
of the tree nodes, as if it was sampled from $\mathcal{P}^+$. This
root sampling method was originally introduced in the POMCP algorithm
\cite{Silver:2010:aab}, developed to solve Partially
Observable MDPs. 

\subsection{BA-UCT with Root Sampling}

The root node of the search tree at a decision point represents the
current state of the BAMDP. The tree is composed of state nodes
representing belief states $\langle s, h \rangle$ and action nodes
representing the effect of particular actions from their parent state node.
The visit counts: $N(\langle s, h\rangle)$ for state nodes, and $N(\langle
s, h \rangle, a)$ for action nodes, are initialized to $0$ and updated
throughout search. A value $Q(\langle s, h \rangle, a)$, initialized to $0$, is also
maintained for each action node. Each simulation traverses the tree 
without backtracking by following the UCT policy at state nodes defined by 
$\argmax_a Q( \langle s, h \rangle, a) + c \sqrt{\sfrac{\log(N(\langle s, h
\rangle))}{N(\langle s, h \rangle, a)}}$, where $c$
is an exploration constant that needs to be set appropriately. Given
an action, the transition distribution $\mathcal{P}^i$
corresponding to the current simulation $i$ is used to sample the next
state. That is, at action node $(\langle s, h
\rangle, a)$, $s'$ is sampled from $\mathcal{P}^i(s,a,\cdot)$, and the new
state node is set to $\langle s', has' \rangle$. When a simulation
reaches a leaf, the tree is expanded by attaching a new state node
with its connected action nodes, and a
rollout policy $\pi_{ro}$ is used to control the MDP defined by the
current $\mathcal{P}^i$ to some fixed depth (determined using the
discount factor). The rollout provides an estimate of the value $Q(\langle s, h \rangle, a)$ from the
leaf action node. This estimate is then used to update the value of all action nodes
traversed during the simulation: if $R$ is the sampled discounted return obtained
from a traversed action node ($\langle s, h \rangle, a$) in a given simulation, then 
we update the value of the action node to 
$Q(\langle s, h \rangle, a) + \sfrac{R-Q(\langle s, h \rangle,a)}{N(\langle s, h \rangle, a)}$ (i.e., the mean
of the sampled returns obtained from that action node over the simulations).
 A detailed description of the BAMCP
algorithm is provided in Algorithm \ref{alg:bmcp}. A diagram example of
BAMCP simulations is presented in Figure \ref{fig:bmcpdi}.

The tree policy treats the forward search as a meta-exploration
problem, preferring to exploit regions of the tree that currently appear better
than others while continuing to explore unknown or less known parts
of the tree. This leads to good empirical results even for small
number of simulations, because effort is expended where search
seems fruitful. Nevertheless all parts of the tree
are eventually visited infinitely often, and therefore the algorithm will
eventually converge on the Bayes-optimal policy (see Section \ref{sec:thprop}).

\begin{figure}[htpb]
\begin{minipage}[b]{0.5\linewidth}
\centering
\begin{algorithm2e}[H]
\caption{BAMCP}
\label{alg:bmcp}
%\dontprintsemicolon
\DontPrintSemicolon
$ $\;
$ $\;
%\Indm
\KwSty{procedure} \FuncSty{Search(} $\langle s,h \rangle$ \FuncSty{)} \; 
\Indp \Repeat{\FuncSty{Timeout()}}{
$\mathcal{P} \sim P(\mathcal{P} | h)$ \;
\FuncSty{Simulate($\langle s,h \rangle, \mathcal{P}, 0$)}\;}
\Return $\displaystyle\argmax_a \; Q( \langle s, h \rangle, a)$\;
\Indm
\KwSty{end procedure}\;
$ $\; 
$ $\; 
\KwSty{procedure} \FuncSty{Rollout(}\!\!\!
                  $\langle s, h \rangle, \mathcal{P}, d$ \FuncSty{)} \;
\Indp \If{$\gamma^{d} Rmax < \epsilon$}{
	\Return $0$\;
}
$a \sim \pi_{ro}(\langle s, h \rangle, \cdot)$\;
$s' \sim \mathcal{P}(s,a,\cdot)$\;
$r \leftarrow \mathcal{R}(s,a)$\;
\Return $r\!+\!\gamma$\FuncSty{Rollout(}$\langle s', has'\rangle, \mathcal{P},d\!+\!1$\FuncSty{)} \;
\Indm
\KwSty{end procedure}\;
$ $\; 
$ $\; 
$ $\; 
\end{algorithm2e}
\end{minipage}
\hspace{0.5cm}
\begin{minipage}[b]{0.5\linewidth}
\centering
\RestyleAlgo{tworuled}
\begin{algorithm2e}[H]
\DontPrintSemicolon
%\Indm
\KwSty{procedure} \FuncSty{Simulate(}
                  $\langle s, h \rangle, \mathcal{P}, d$\FuncSty{)} \;
\Indp
\lIf{$\gamma^{d} Rmax < \epsilon$}{
	\Return $0$\;
}
\If{$N(\langle s,h \rangle) = 0$}{
	\For{\KwSty{all} $a \in A$}{
		$N(\langle s, h \rangle, a) \leftarrow 0$, $Q(\langle s, h \rangle,a)) \leftarrow 0$\;
	}
	$a \sim \pi_{ro}(\langle s,h\rangle, \cdot)$\;
	$s' \sim \mathcal{P}(s,a,\cdot)$\;
	$r \leftarrow \mathcal{R}(s,a)$\;
	$R \leftarrow r + \gamma$ \FuncSty{Rollout(}$\langle s',has'\rangle, \mathcal{P}, d$\FuncSty{)}\;	
	$N(\langle s, h \rangle) \leftarrow 1$, $N(\langle s, h \rangle, a) \leftarrow 1$\;
	$Q(\langle s, h \rangle, a) \leftarrow R$\;
	\Return  $R$\;
}
$a \leftarrow \displaystyle\argmax_b Q( \langle s, h \rangle, b) + c \sqrt{\tfrac{\log(N(\langle s, h \rangle))}{N(\langle s, h \rangle, b)}}$\;
$s' \sim \mathcal{P}(s,a,\cdot)$\;
$r \leftarrow \mathcal{R}(s,a)$\;
$R \leftarrow r + \gamma $ \FuncSty{Simulate(}$\langle s', has'\rangle, \mathcal{P}, d\!+\!1$\FuncSty{)} \;
$N(\langle s, h \rangle) \leftarrow N(\langle s, h \rangle) + 1$\;
$N(\langle s, h \rangle, a) \leftarrow N(\langle s, h \rangle, a) + 1$\;
$Q(\langle s, h \rangle, a) \leftarrow Q(\langle s, h \rangle, a) + \frac{R-Q(\langle s, h \rangle,a)}{N(\langle s, h \rangle, a)}$\;
\Return $R$\;
\Indm
\KwSty{end procedure}\;
\end{algorithm2e}
\end{minipage}
\end{figure}

Finally, note that the history of transitions $h$ is generally not the
most compact sufficient statistic of the belief in fully observable
MDPs. Indeed, it can be replaced with unordered transition counts
$\psi$, considerably reducing the number of states of the BAMDP and,
potentially the complexity of planning.  Given an addressing scheme
suitable to the resulting expanding lattice (rather than to a tree),
BAMCP can search in this reduced space. We found this version of BAMCP
to offer only a marginal improvement. This is a common finding for
UCT, stemming from its tendency to concentrate search effort on one of
several equivalent paths (up to transposition), implying a limited
effect on performance of reducing the number of those paths.

\subsection{Lazy Sampling}
\label{sec:lazysamp}

In previous work on sample-based tree search, indeed including
POMCP~\cite{Silver:2010:aab}, a complete sample state is drawn from the
posterior at the root of the search tree.  However, this can be
computationally very costly. Instead, we sample $\mathcal{P}$ lazily,
creating only the particular transition probabilities that are
required as the simulation traverses the tree, and also during the
rollout.

Consider $\mathcal{P}(s,a,\cdot)$ to be parametrized by a latent
variable $\theta_{s,a}$ for each state and action pair. These may
depend on each other, as well as on an additional set of latent
variables $\phi$. The posterior over $\mathcal{P}$ can be written as
$P(\Theta | h) = \int_{\phi} P(\Theta | \phi , h) P(\phi | h)$, where $\Theta = \{ \theta_{s,a} | s \in S, a \in A \}$.  
Define $\Theta_t = \{ \theta_{s_1,a_1}, \cdots, \theta_{s_t,a_t} \}$
as the (random) set of $\theta$ parameters required 
during the course of a BAMCP simulation that starts at time $1$ and ends at time $t$.
Using the chain rule, we can rewrite 
\begin{align*}
P(\Theta | \phi, h) = P(\theta_{s_1,a_1} | \phi, h)  P(\theta_{s_2,a_2} | \Theta_1, \phi, h)	
  \dots  
 P(\theta_{s_T,a_T} | \Theta_{T-1}, \phi, h)   P(\Theta \setminus \Theta_T | \Theta_T, \phi, h)  
\end{align*} 
where $T$ is the length of the simulation and $\Theta \setminus \Theta_T$ denotes the (random) set 
of parameters that are not required for a simulation. 
For each simulation $i$, we sample $P(\phi | h_t)$ at the root and then lazily sample the $\theta_{s_t,a_t}$ parameters as required, conditioned 
on $\phi$ and all $\Theta_{t-1}$ parameters sampled for the current simulation. This process is stopped at the end of the simulation, potentially
before all $\theta$ parameters have been sampled. For example, if the
transition parameters for different states and actions are 
independent, we can completely forgo sampling a complete $\mathcal{P}$, and instead
draw any necessary parameters individually for each state-action pair. This leads to substantial performance improvement, especially
in large MDPs where a single simulation only requires a small subset of parameters (see for example the domain in Section~\ref{sec:infgrid}).

\subsection{Rollout Policy Learning}
\label{sec:ro}

The choice of rollout policy $\pi_{ro}$ is important if simulations
are few, especially if the domain does not display substantial locality
or if rewards require a carefully selected sequence of actions to be
obtained. Otherwise, a simple uniform random policy can be chosen to
provide noisy estimates. 
In this work, we learn $Q_{ro}$, the optimal $Q$-value in the real MDP, 
in a model-free manner (e.g., using Q-learning) from samples $(s_t,a_t,r_t,s_{t+1})$ 
obtained off-policy as a result of the interaction of the Bayesian agent with the environment. 
Acting greedily according to $Q_{ro}$ translates to pure
exploitation of gathered knowledge. A rollout policy in BAMCP following $Q_{ro}$ could
therefore over-exploit. Instead,   
similar to \cite{Gelly:2007:aaa}, we select an $\epsilon$-greedy policy with 
respect to $Q_{ro}$ as our rollout policy $\pi_{ro}$. This biases rollouts towards observed regions 
of high rewards. This method provides valuable direction for the rollout policy at negligible
computational cost. More complex rollout policies can be considered, for example rollout policies
that depend on the sampled model $\mathcal{P}^i$. However, these
usually incur computational overhead. 

\subsection{Theoretical properties}
\label{sec:thprop}

Define $V(\langle s, h \rangle) = \displaystyle\max_{a \in A} Q(\langle s, h \rangle, a) \;\; \forall \langle s,h \rangle \in S \times \mathcal{H}$.

\newtheorem{th:convergence}{Theorem}
\begin{th:convergence}
For all $\epsilon > 0$ (the numerical precision, see Algorithm~\ref{alg:bmcp}) and a suitably chosen $c$ (e.g. $c >
\frac{Rmax}{1-\gamma}$), from state $\langle s_t, h_t \rangle$, BAMCP constructs
a value function at the root node that converges in probability to an $\epsilon'$-optimal value
function, $V(\langle s_t, h_t \rangle) \overset{p}{\rightarrow}
V^*_{\epsilon'}(\langle s_t, h_t \rangle)$, where $\epsilon' = \frac{\epsilon}{1-\gamma}$.
Moreover, for large enough $N(\langle s_t,h_t \rangle)$, the bias of
$V(\langle s_t, h_t \rangle)$ decreases as $O(\log(N(\langle s_t, h_t \rangle)) /
N(\langle s_t, h_t \rangle))$. \text{(Proof available in supplementary material)}
\label{th:conv}
\end{th:convergence} 

By definition, Theorem \ref{th:conv} implies that BAMCP converges to the
Bayes-optimal solution asymptotically. We confirmed this result
empirically using a variety of Bandit problems, for which the Bayes-optimal
solution can be computed efficiently using Gittins indices (see supplementary material).

%%%%%%%%%%%%%%%%%%%%%%%%%%%%%%%%%%%%%%%%%%%
%
%   RELATED WORK
%
%
%%%%%%%%%%%%%%%%%%%%%%%%%%%%%%%%%%%%%%%%%%%
\vspace{-0.16in}
\section{Related Work}
\label{sec:relwork}

In Section~\ref{sec:exp}, we compare BAMCP to a set of existing
Bayesian RL algorithms. Given limited space, we do not provide a
comprehensive list of planning algorithms for MDP exploration, but
rather concentrate on related
sample-based algorithms for Bayesian RL.

Bayesian DP \cite{Strens:2000:aaa} maintains a posterior distribution over
transition models. At each step, a single model is sampled, and the action that
is optimal in that model is executed. The Best Of Sampled Set (BOSS) algorithm
generalizes this idea \cite{Asmuth:2009:aaa}.  BOSS samples a number of models
from the posterior and combines them optimistically. This drives sufficient
exploration to guarantee finite-sample performance guarantees. BOSS is quite
sensitive to its parameter that governs the sampling criterion. Unfortunately,
this is difficult to select. Castro and Precup  proposed an SBOSS variant,
which provides a more effective adaptive sampling
criterion~\cite{Castro:2010:aaa}. BOSS algorithms are generally
quite robust, but suffer from over-exploration.

%\subsection{Forward-Search Methods}

Sparse sampling~\cite{Kearns:1999:aaa} is a sample-based tree search algorithm.
The key idea is to sample successor nodes from each state, and apply a Bellman
backup to update the value of the parent node from the values of the child
nodes. Wang et al. applied sparse sampling to search over belief-state
MDPs\cite{Wang:2005:aaa}.  The tree is expanded non-uniformly
according to the sampled trajectories. At each decision node, a
promising action is selected 
using Thompson sampling --- i.e., sampling an MDP from that belief-state,
solving the MDP and taking the optimal action. At each chance node, a successor
belief-state is sampled from the transition dynamics of the belief-state MDP.

Asmuth and Littman further extended this idea in their BFS3
algorithm~\cite{Asmuth:2011:aaa}, an adaptation of Forward Search Sparse
Sampling \cite{Walsh:2010:aaa} to belief-MDPs. Although they described their
algorithm as Monte-Carlo tree search, it in fact uses a Bellman backup rather
than Monte-Carlo evaluation. Each Bellman backup updates both lower and upper
bounds on the value of each node. Like Wang et al., the tree is expanded
non-uniformly according to the sampled trajectories, albeit using a
different method for action selection. At each decision node, a
promising action is 
selected by maximising the upper bound on value. At each chance node,
observations are selected by maximising the uncertainty (upper minus lower
bound).

Bayesian Exploration Bonus (BEB) solves the posterior mean MDP, but with an
additional reward bonus that depends on visitation counts
\cite{Kolter:2009:aaa}. Similarly, Sorg et al. propose an algorithm with a
different form of exploration bonus~\cite{Sorg:2010:aac}.  These algorithms
provide performance guarantees after a polynomial number of steps in the
environment. However, behavior in the early steps of exploration is very
sensitive to the precise exploration bonuses; and it turns out to be hard to
translate sophisticated prior knowledge into the form of a bonus.

\begin{table*}[th]
\vspace{-0.12in}
\caption{\small{Experiment results summary. For each algorithm, we report the
mean sum of rewards and confidence interval for the best performing parameter
within a reasonable planning time limit (0.25 s/step for Double-loop, 1 s/step
for Grid5 and Grid10, 1.5 s/step for the Maze). For BAMCP, this simply
corresponds to the number of simulations that achieve a planning time just
under the imposed limit. * Results reported from \cite{Strens:2000:aaa} without
timing information.}} 
\label{tab:results}
\small{
\begin{center}
    \begin{tabular}{|l|c|c|c|c|}
        \hline
        ~	& Double-loop &  Grid5 & Grid10 & Dearden's Maze \\ \hline
        BAMCP & \textbf{387.6 $\pm$ 1.5}       &    \textbf{72.9 $\pm$ 3}  & \textbf{32.7 $\pm$ 3}   & \textbf{965.2 $\pm$ 73} \\ 
        BFS3 \cite{Asmuth:2011:aaa}	& 382.2 $\pm$ 1.5 &    66 $\pm$ 5 & 10.4 $\pm$ 2   & 240.9 $\pm$ 46   \\ 
        SBOSS \cite{Castro:2010:aaa}	& 371.5 $\pm$ 3    &  59.3 $\pm$ 4 & 21.8 $\pm$ 2   & 671.3 $\pm$ 126 \\ 
        BEB \cite{Kolter:2009:aaa}							& 386 $\pm$ 0      &   67.5 $\pm$ 3  & 10 $\pm$ 1 & 184.6 $\pm$ 35 \\ 
        Bayesian DP* \cite{Strens:2000:aaa}			& 377 $\pm$ 1       		& -        & -        & -            \\ 
        Bayes VPI+MIX* \cite{Dearden:1998:aaa}	& 326 $\pm$ 31  		& -        & -        & 817.6  $\pm$ 29  \\ 
        IEQL+* \cite{Meuleau:1999:aaa}					& 264 $\pm$ 1   			& -        & -        & 269.4 $\pm$ 1      \\ 
        QL Boltzmann*														& 186 $\pm$ 1   		& -        & -        & 195.2 $\pm$ 20   \\ 
        \hline
    \end{tabular}
	\end{center}}
\end{table*}
\vspace{-0.1in}

%%%%%%%%%%%%%%%%%%%%%%%%%%%%%%%%%%%%%%%%%%%
%
%   EXPERIMENTS
%
%
%%%%%%%%%%%%%%%%%%%%%%%%%%%%%%%%%%%%%%%%%%%
\section{Experiments}
\label{sec:exp}
\vspace{-0.1in}

We first present empirical results of BAMCP on a set of standard problems with comparisons 
to other popular algorithms. Then we showcase BAMCP's advantages in a
large scale 
task: an infinite 2D grid with complex correlations between reward locations.

\vspace{-0.09in}
\subsection{Standard Domains}
\vspace{-0.04in}
\paragraph{Algorithms}
\label{sec:algs}
\vspace{-0.1in}

The following algorithms were run:
\textbf{BAMCP} - The algorithm presented in Section \ref{sec:bmcp},
implemented with lazy sampling. The algorithm was run for different
number of simulations (10 to 10000) to span different planning
times. In all experiments, we set $\pi_\text{ro}$ to be an
$\epsilon$-greedy policy with $\epsilon=0.5$. The UCT exploration
constant was left unchanged for all experiments $(c=3)$, we
experimented with other values of $c \in \{0.5, 1, 5 \}$ with similar
results. 
\textbf{SBOSS} \cite{Castro:2010:aaa}: for 
each domain, we varied the number of samples
$K \in \{2,4,8,16,32\}$ and the resampling threshold parameter $\delta
\in \{3,5,7\}$. 
\textbf{BEB} \cite{Kolter:2009:aaa}: for each domain, we varied the
bonus parameter $\beta \in \{0.5,1,1.5,2,2.5,3,5,10,15,20\}$. 
\textbf{BFS3} \cite{Asmuth:2011:aaa} for each domain, we varied the branching
factor $C \in \{2,5,10,15\}$ and the number of simulations (10 to
2000). The depth of search was set to $15$ in all domains except for
the larger grid and maze domain where it was set to $50$. We also
tuned the $V_{\text{max}}$ parameter for each domain ---
$V_{\text{min}}$ was always set to $0$. 
In addition, we report results from~\cite{Strens:2000:aaa} for several other prior algorithms.

\vspace{-0.2in}
\paragraph{Domains}
\vspace{-0.1in}

For all domains, we fix $\gamma=0.95$.
%\begin{itemize}
The \textbf{Double-loop} domain is a $9$-state deterministic MDP with
$2$ actions \cite{Dearden:1998:aaa}, $1000$ steps are executed in this
domain.  
\textbf{Grid5} is a $5\times5$ grid with no reward anywhere except for a reward
state opposite to the reset state. Actions with cardinal directions are
executed with small probability of failure for $1000$ steps.
\textbf{Grid10} is a $10\times10$ grid designed like Grid5. We collect $2000$
steps in this domain.
\textbf{Dearden's Maze} is a $264$-states maze with $3$ flags to collect
\cite{Dearden:1998:aaa}. A special reward state gives the number of flags
collected since the last visit as reward, $20000$ steps are executed in this
domain.~\footnote{The result reported for Dearden's maze with the Bayesian DP
alg. in \cite{Strens:2000:aaa} is for a different version of the task in which
the maze layout is given to the agent.} 

%\end{itemize}

To quantify the performance of each algorithm, we measured the total
undiscounted reward over many steps. We chose this measure of performance to
enable fair comparisons to be drawn with prior work. In fact, we are optimising
a different criterion -- the discounted reward from the start state -- and so
we might expect this evaluation to be unfavourable to our algorithm. 

One major advantage of Bayesian RL is that one can specify priors about the
dynamics.  For the Double-loop domain, the Bayesian RL algorithms were run with
a simple Dirichlet-Multinomial model with symmetric Dirichlet parameter
$\alpha=\frac{1}{|S|}$. For the grids and the maze domain, the algorithms were
run with a sparse Dirichlet-Multinomial model, as described in
\cite{Friedman:1999:aaa}. For both of these models, efficient collapsed
sampling schemes are available; they are employed for the BA-UCT and BFS3
algorithms in our experiments to compress the posterior parameter sampling and
the transition sampling into a single transition sampling step. This
considerably reduces the cost of belief updates inside the search tree when
using these simple probabilistic models. In general, efficient collapsed
sampling schemes are not available (see for example the model in
Section~\ref{sec:infgrid}).

\begin{figure}[th]
\centering
\subfigure{
(a) \includegraphics[width=0.88\textwidth]{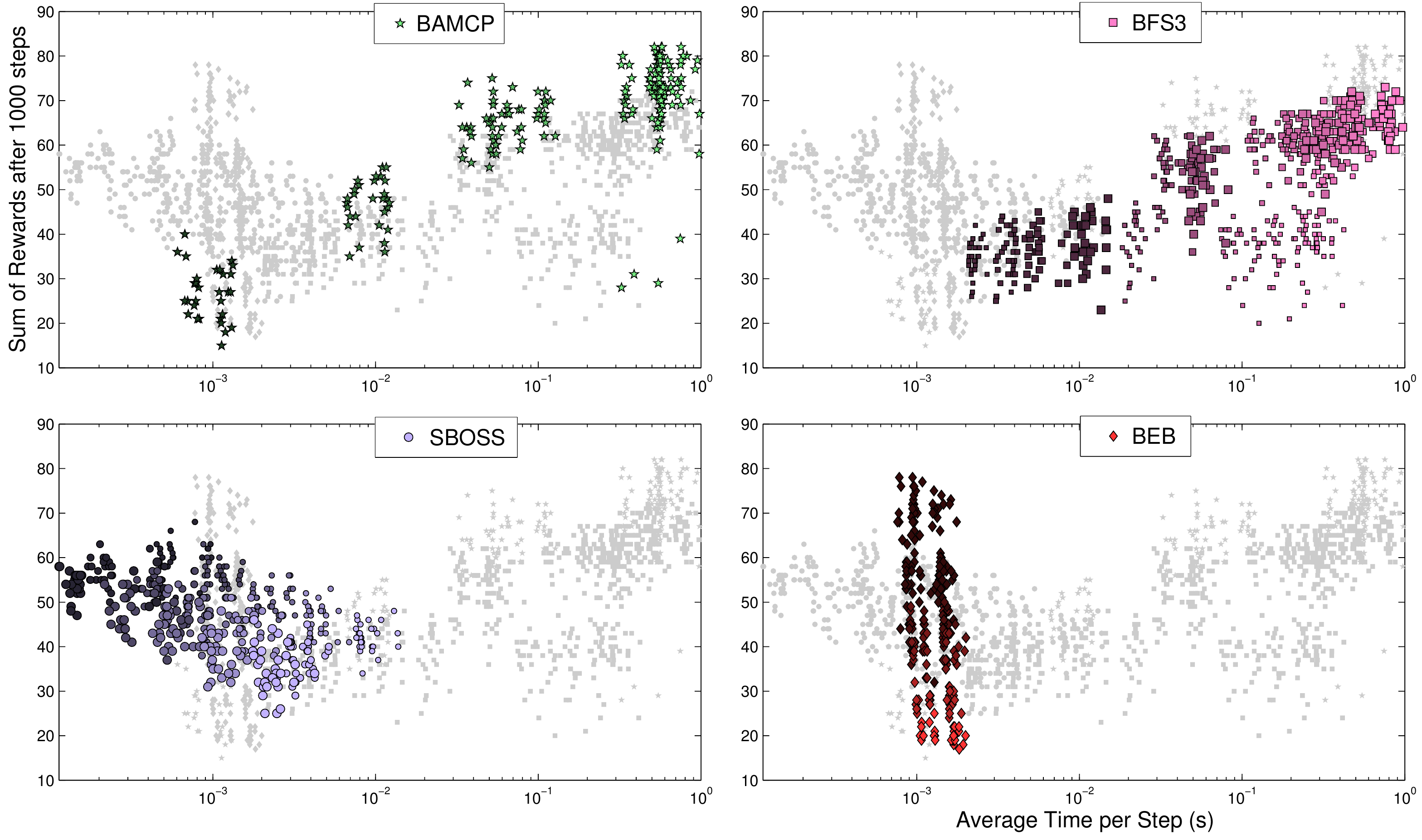}
\label{fig:grid5}
}
\subfigure[]{
\includegraphics[width=0.33\textwidth]{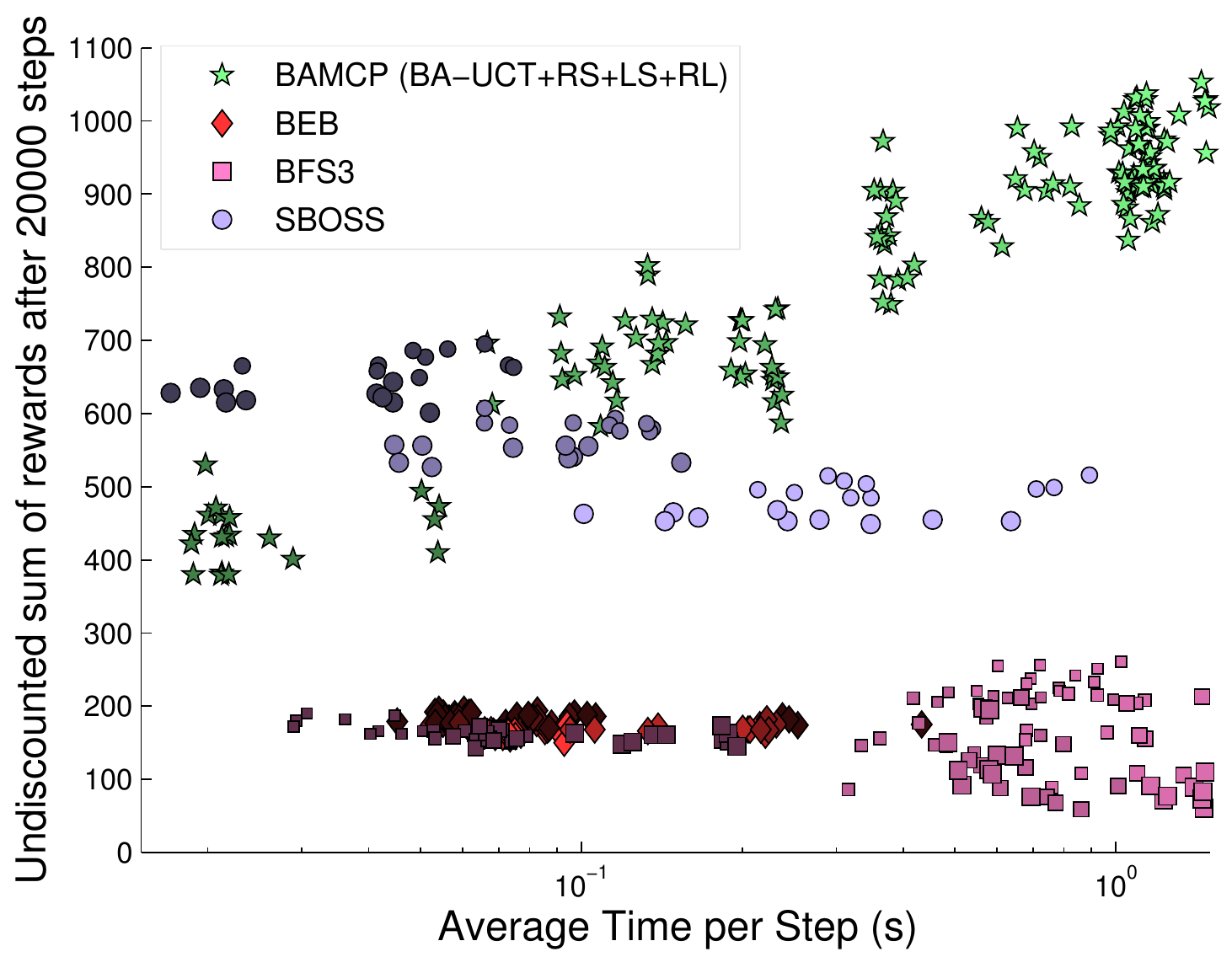}
\label{fig:maze1}
}
\subfigure[]{
\includegraphics[width=0.3\textwidth]{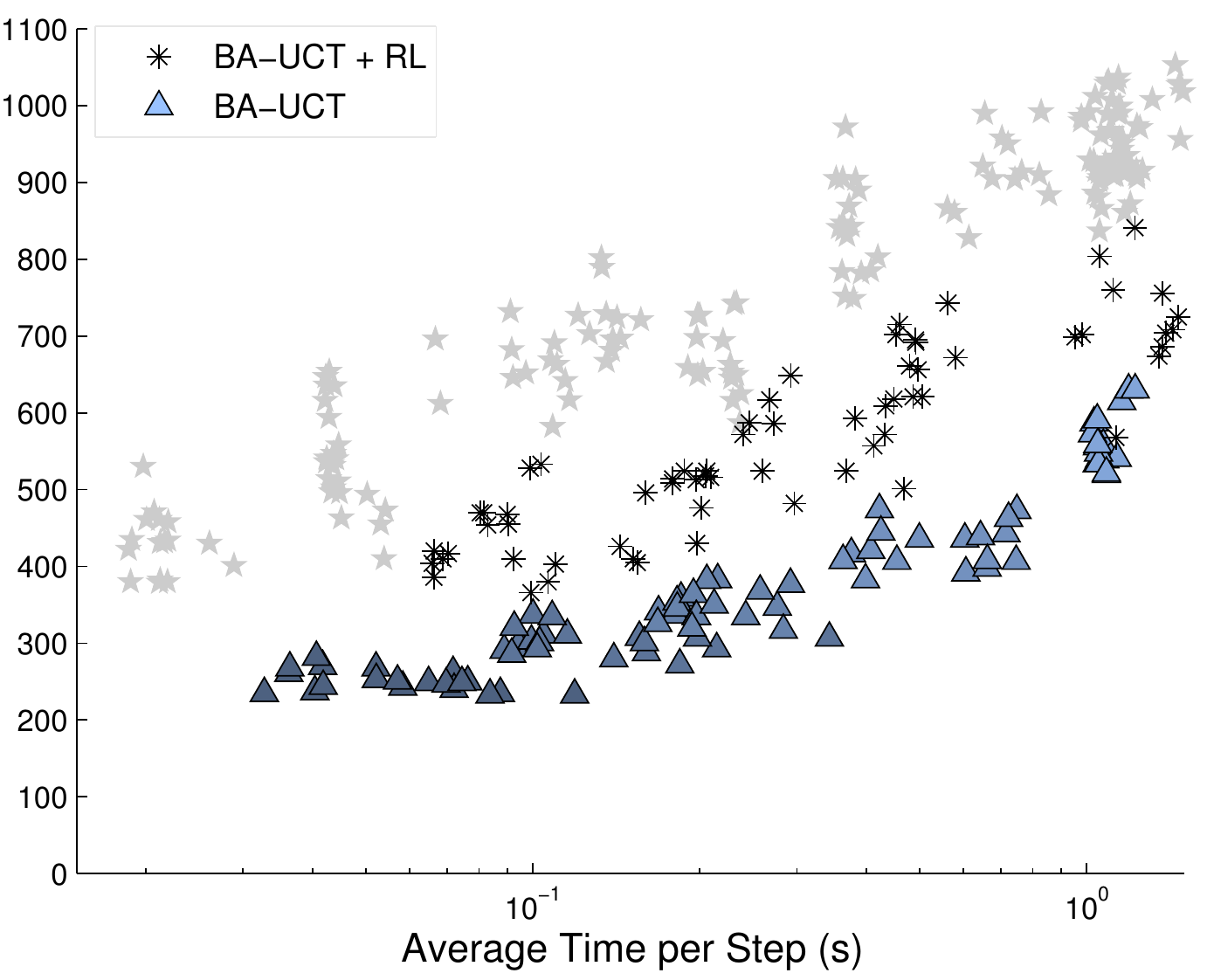}
\label{fig:maze2}
}
\subfigure[]{
\includegraphics[width=0.3\textwidth]{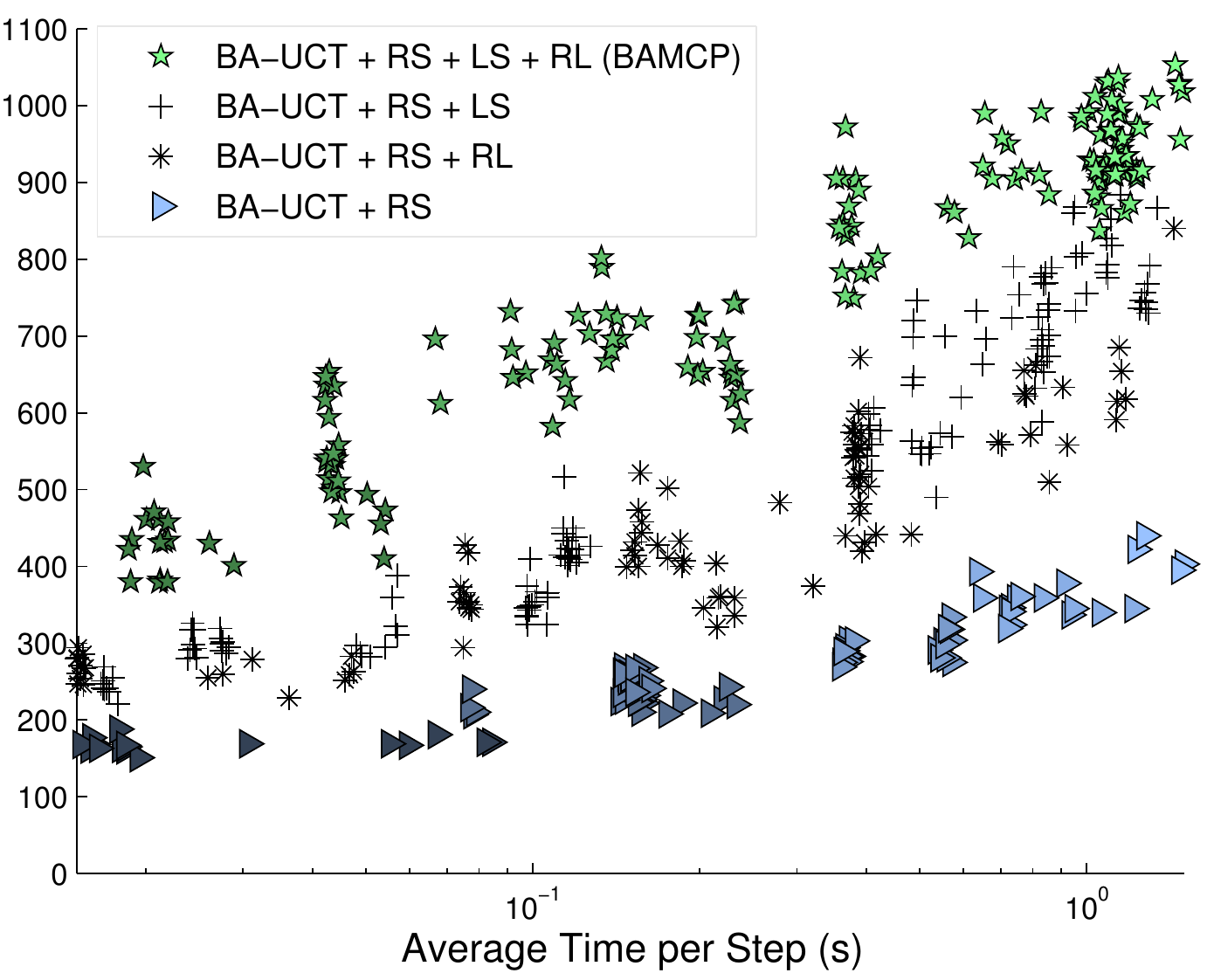}
\label{fig:maze3}
}
\vspace{-0.08in}
\caption{\small{
Performance of each algorithm on the Grid5 (\textbf{a.}) and Maze domain
(\textbf{b-d}) as a function of planning time. Each point corresponds to a
single run of an algorithm with an associated setting of the parameters.
Increasing brightness inside the points codes for an increasing value of a
parameter (BAMCP and BFS3: number of simulations, BEB: bonus parameter $\beta$,
SBOSS: number of samples $K$). A second dimension of variation is coded as the
size of the points (BFS3: branching factor $C$, SBOSS: resampling parameter
$\delta$). The range of parameters is specified in Section~\ref{sec:algs}.
\textbf{a.} Performance of each algorithm on the Grid5 domain.
\textbf{b.} Performance of each algorithm on the Maze domain.
\textbf{c.} On the Maze domain, performance of vanilla BA-UCT with and without rollout policy learning (RL).
\textbf{d.} On the Maze domain, performance of BAMCP with and without the lazy sampling (LS) and rollout
policy learning (RL) presented in Sections~\ref{sec:ro}, \ref{sec:lazysamp}. Root sampling (RS) is included.}}
\label{fig:gridmaze}
\vspace{-0.1in}
\end{figure}

\vspace{-0.18in}
\paragraph{Results}
\vspace{-0.08in}

A summary of the results is presented in Table \ref{tab:results}. Figure~\ref{fig:gridmaze} reports
 the planning time/performance trade-off for the different algorithms on the Grid5 and Maze domain.

On all the domains tested, BAMCP performed best. 
Other algorithms came close on some tasks, but only when their parameters were
tuned to that specific domain. This is particularly evident for BEB, which
required a different value of exploration bonus to achieve maximum performance
in each domain. BAMCP's performance is stable with respect to the choice of its
exploration constant $c$ and it did not require tuning to obtain the results.

BAMCP's performance scales well as a function of planning time, as is evident in
Figure~\ref{fig:gridmaze}. In contrast, SBOSS follows the
opposite trend. If more samples are employed to build the merged model, SBOSS
actually becomes too optimistic and over-explores,
degrading its performance. BEB cannot take advantage of prolonged planning time at all. BFS3
generally scales up with more planning time with an appropriate choice
of parameters, but it is not obvious how to trade-off the branching
factor, depth, and number of simulations in each domain.
BAMCP greatly benefited from our lazy sampling scheme in the experiments, providing $35\times$ speed
improvement over the naive approach in the maze domain for example; this is illustrated in
Figure~\ref{fig:maze2}.

Dearden's maze aptly illustrates a major drawback of forward search sparse
sampling algorithms such as BFS3. Like many maze problems, all rewards are zero
for at least $k$ steps, where $k$ is the solution length. Without prior
knowledge of the optimal solution length, all upper bounds will be higher than
the true optimal value until the tree has been fully expanded up to depth $k$
-- even if a simulation happens to solve the maze. In contrast, once BAMCP
discovers a successful simulation, its Monte-Carlo evaluation will immediately
bias the search tree towards the successful trajectory.

\begin{figure*}[th]
\begin{center}
	\includegraphics[width=0.6\textwidth]{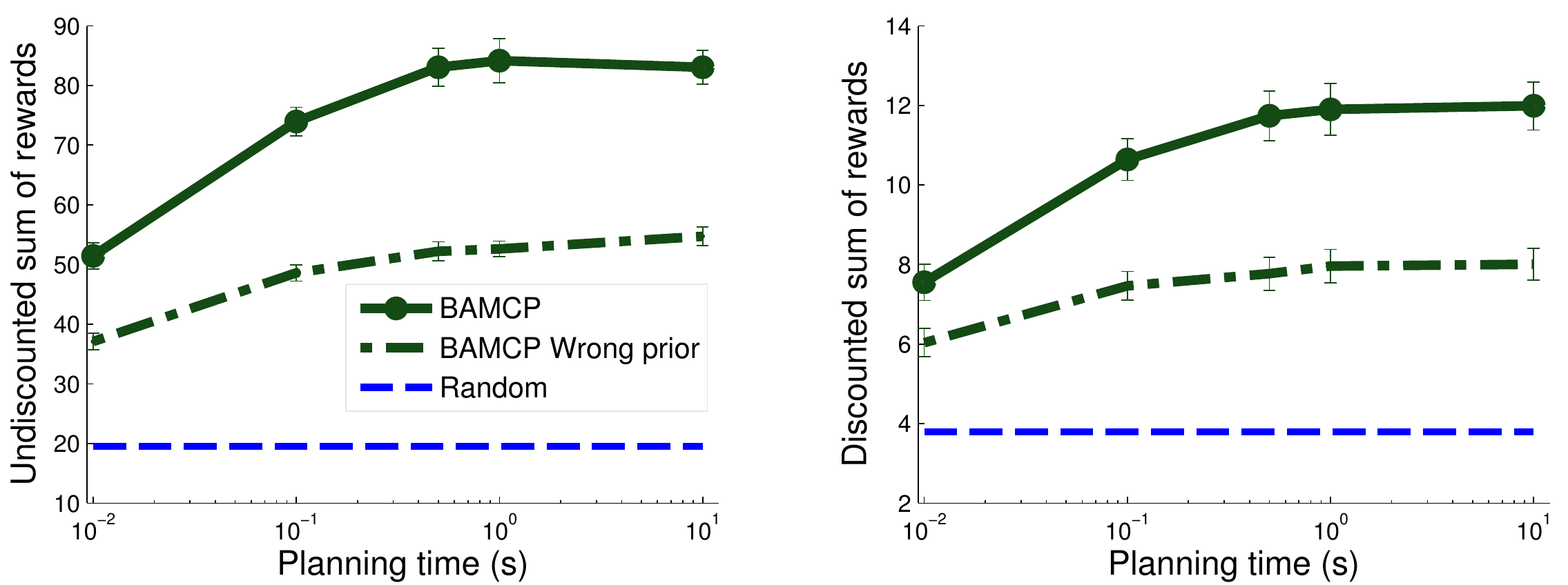}
  \vspace{-0.05in}
	\caption{\small{Performance of BAMCP as a function of planning time on the
	Infinite 2D grid task of Section~\ref{sec:infgrid}, for $\gamma=0.97$, where
	the grids are generated with Beta parameters
	$\alpha_1=1,\beta_1=2,\alpha_2=2,\beta_2=1$ (See supp.
	Figure~\ref{fig:infgridviz} for a visualization).  The performance during the
	first 200 steps in the environment is averaged over 50 sampled environments
	(5 runs for each sample) and is reported both in terms of undiscounted (left)
	and discounted (right) sum of rewards. BAMCP is run either with the correct
	generative model as prior or with an incorrect prior (parameters for rows and
	columns are swapped), it is clear that BAMCP can take advantage of correct
	prior information to gain more rewards. The performance of a uniform random
	policy is also reported.
}}
\label{fig:infgridset3}
\end{center}
\vspace{-0.12in}
\end{figure*}

\vspace{-0.09in}
\subsection{Infinite 2D grid task}
\label{sec:infgrid}
\vspace{-0.02in}

We also applied BAMCP to a much larger problem. The generative model for this infinite-grid MDP is as follows: each column $i$
has an associated latent parameter $p_i \sim \text{Beta}(\alpha_1,\beta_1)$ and each
row $j$ has an associated latent parameter $q_j \sim \text{Beta}(\alpha_2,
\beta_2)$. The probability of grid cell $ij$ having a reward of $1$ is
$p_iq_j$, otherwise 
the reward is $0$. The agent knows it is on a grid and is always free to move in any of the four cardinal
directions. Rewards are consumed when visited; returning to the same location subsequently results in a reward of 0. As opposed
to the independent Dirichlet priors employed in standard domains,
here, dynamics
are tightly correlated across states  (i.e., observing a state
transition provides information about other state transitions).
Posterior inference (of the dynamics $\mP$) in this model requires
approximation because of the non-conjugate coupling of the variables, the
inference  is done via MCMC (details in Supplementary). The domain is illustrated in
Figure~\ref{fig:infgridviz}.

Planning algorithms that attempt to solve an MDP based on sample(s) (or the
mean) of the posterior (e.g., BOSS, BEB, Bayesian DP) cannot directly handle the large
state space. Prior forward-search methods (e.g., BA-UCT, BFS3) can deal with the state space, but not the large belief space: at every node of the search tree they must solve an approximate inference problem to estimate the posterior beliefs. In contrast, BAMCP limits the posterior
inference to the root of the search tree and is not directly affected by the
size of the state space or belief space, which allows the algorithm to perform well even with a
limited planning time. Note that lazy sampling is required in this setup since
a full sample of the dynamics involves infinitely many parameters.

Figure~\ref{fig:infgridset3} (and Figure~\ref{fig:infgridset12})
demonstrates the planning performance of BAMCP in this complex domain.
Performance improves with additional planning time, and the quality of the
prior clearly affects the agent's performance. Supplementary videos 
contrast the behavior of the agent for different prior
parameters.

%%%%%%%%%%%%%%%%%%%%%%%%%%%%%%%%%%%%%%%%%%%
%
%   FUTURE WORK
%
%
%%%%%%%%%%%%%%%%%%%%%%%%%%%%%%%%%%%%%%%%%%%
\vspace{-0.12in}
\section{Future Work}
\vspace{-0.07in}

The UCT algorithm is known to have several drawbacks. First, there are no
finite-time regret bounds. It is possible to construct malicious environments,
for example in which the optimal policy is hidden in a generally low reward
region of the tree, where UCT can be misled for long
periods~\cite{Coquelin:2007:aaa}. Second, the UCT algorithm treats every action
node as a multi-armed bandit problem. However, there is no actual benefit to
accruing reward during planning, and so it is in theory more appropriate to use
\emph{pure exploration} bandits~\cite{Bubeck:2009:aaa}. Nevertheless, the UCT
algorithm has produced excellent empirical performance in many domains \cite{Gelly:2012:aaa}.

BAMCP is able to exploit prior knowledge about the dynamics in a principled
manner. In principle, it is possible to encode many aspects of domain knowledge
into the prior distribution. An important avenue for future work is to explore
rich, structured priors about the dynamics of the MDP. If this prior knowledge
matches the class of environments that the agent will encounter, then
exploration could be significantly accelerated.

%%%%%%%%%%%%%%%%%%%%%%%%%%%%%%%%%%%%%%%%%%%
%
%   Conclusion
%
%%%%%%%%%%%%%%%%%%%%%%%%%%%%%%%%%%%%%%%%%%%
\vspace{-0.12in}
\section{Conclusion}
\vspace{-0.07in}

We suggested a sample-based algorithm for Bayesian RL called BAMCP that
significantly surpassed the performance of existing algorithms on several
standard tasks. We showed that BAMCP can tackle larger and more
complex tasks generated from a structured prior, where existing
approaches scale poorly. In addition, BAMCP provably converges to the
Bayes-optimal solution.

The main idea is to employ Monte-Carlo tree search to explore the augmented
Bayes-adaptive search space efficiently. The naive implementation of that idea
is the proposed BA-UCT algorithm, which cannot scale for most priors due to
expensive belief updates inside the search tree. We introduced three
modifications to obtain a computationally tractable sample-based algorithm:
root sampling, which only requires beliefs to be sampled at the start of
each simulation (as in \cite{Silver:2010:aab}); a model-free
RL algorithm that learns a rollout policy; and the
use of a lazy sampling scheme to sample the posterior beliefs cheaply.

%\subsubsection*{Acknowledgements}

\clearpage
\bibliographystyle{plain}
\begin{footnotesize}
	\bibliography{arxiv_bamcp}
\end{footnotesize}

\clearpage
\setcounter{figure}{0}
\makeatletter 
\renewcommand{\thefigure}{S\@arabic\c@figure} 
{\Large \textbf{Supplementary Material}}

\paragraph{ {\large Proof of Theorem~\ref{th:conv} and comments}}

Consider the BA-UCT algorithm: UCT applied to the Bayes-Adaptive MDP (dynamics are described
in Equation~\ref{eq:BAMDP}). Let $\mD^\pi(h_T)$ be the \emph{
rollout distribution} of BA-UCT: the probability that history $h_T$ is generated when
running the BA-UCT search from $\langle s_t, h_t \rangle$, with $h_t$ a prefix
of $h_T$, $T-t$ the effective horizon in the search tree, and $\pi$ an
arbitrary BAMDP policy. Similarly define the similar quantity $\tilde{\mD}^\pi(h_T)$:
the probability that history $h_T$ is generated when running the BAMCP algorithm. The following
lemma shows that these two quantities are in fact equivalent.\footnote{For ease of notation,
we refer to a node with its history as opposed to its state and history as done in the rest of the paper.}

\newtheorem{th:distribution}{Lemma}
\begin{th:distribution}
	$\mD^\pi(h_T) = \tilde{\mD}^\pi(h_T)$ for all BAMDP policies $\pi: \mathcal{H} \rightarrow A$.
\label{th:distr}
\end{th:distribution}
\begin{proof}
	Let $\pi$ be arbitrary. We show by induction that for all suffix histories $h$ of $h_t$, 
	$\mD^\pi(h) = \tilde{\mD}^\pi(h)$; but also $P(\mP | h) =
	\tilde{P}_h(\mP)$ where $P(\mP | h)$ denotes (as before) the posterior distribution over
	the dynamics given $h$ and $\tilde{P}_h(\mP)$ denotes the distribution
	of $\mP$ at node $h$ when running BAMCP.
	
	\vspace{0.1in}
	\emph{Base case:} At the root ($h = h_t$, suffix history of size 0), it is clear that $\tilde{P}_{h_{t}}(\mP) = P(\mP | h_{t})$ 
	since we are sampling from the posterior at the root node and $D^\pi(h_t)=\tilde{\mD}^\pi(h_t)=1$ since all simulations
	go through the root node.
	
	\vspace{0.1in}
	\emph{Step case:} 
		
	Assume proposition true for all suffices of size $i$. Consider any suffix $has'$ of size $i+1$, where $a \in A$ and $s' \in S$ are 
	arbitrary and $h$ is an arbitrary suffix of size $i$ ending in $s$. The following relation holds: 
	\begin{align}
		\mD^\pi(has') &= \mD^\pi(h) \pi(h,a) \int_{\mP} \mathrm{d}\!\mP P(\mP | h) \mP(s,a,s') \\
		&= \tilde{\mD}^\pi(h) \pi(h,a) \int_{\mP} \mathrm{d}\!\mP \tilde{P}_h(\mP) \mP(s,a,s') \\
		&= \tilde{\mD}^\pi(has'), 
	\end{align}
  where the second line is obtained using the induction hypothesis, and the rest from the definitions.
	In addition, we can match the distribution of the samples $\mP$ at node $has'$:
	\begin{align}
		P(\mP | has') & = P(has' | \mP) P(\mP) / P(has') \\
		             & = P(h | \mP) P(\mP) \mP(s,a,s') / P(has')\\
								 &= P(\mP | h) P(h) \mP(s,a,s') / P(has') \\
                 &= Z P(\mP | h) \mP(s,a,s') \\
								 &= Z \tilde{P}_{h}(\mP) \mP(s,a,s') \label{eq:PIH}\\
								 &= Z \tilde{P}_{ha}(\mP) \mP(s,a,s') \label{eq:PA}\\
								 &= \tilde{P}_{has'}(\mP),
								 \label{eq:PSP}
	\end{align}
	where Equation~\ref{eq:PIH} is obtained from the induction hypothesis,
	Equation~\ref{eq:PA} is obtained from the fact that the choice of action at
	each node is made independently of the samples $\mP$. Finally, to obtain Equation~\ref{eq:PSP}
	from Equation~\ref{eq:PA}, consider the probability that a sample $\mP$ arrives at node 
	$has'$, it first needs to traverse node $ha$ (this occurs with probability $\tilde{P}_{ha}(\mP)$)
	and then, from node $ha$, the state $s'$ needs to be sampled (this occurs with probability $\mP(s,a,s')$);
	therefore, $\tilde{P}_{has'}(\mP) \propto \tilde{P}_{ha}(\mP) \mP(s,a,s')$. $Z$ is the normalization
  constant: $Z=\sfrac{1}{\int_{\mP} \mP(s,a,s') P(\mP|h)}=\sfrac{1}{\int_{\mP} \mP(s,a,s') \tilde{P}_{h}(\mP)}$. This completes the induction.	
\end{proof}

\begin{proof}[Proof of Theorem~\ref{th:conv}]
	The UCT analysis in Kocsis and Szepesv{\'a}ri~\cite{Kocsis:2006:aaa} applies to the
BA-UCT algorithm, since it is 
 vanilla UCT applied to the BAMDP (a particular MDP). By Lemma~\ref{th:distr}, BAMCP
 simulations are equivalent in distribution to BA-UCT simulations. The nodes in BAMCP are therefore
 being evaluated as in BA-UCT, providing the result.
\end{proof}

Lemma~\ref{th:distr} provides some intuition for why belief updates
are unnecessary in 
the search tree: the search tree filters the samples from the root node so that
the distribution of samples at each node is equivalent to the distribution
obtained when explicitly updating the belief. In particular, the root sampling
in POMCP \cite{Silver:2010:aab} and BAMCP is different from evaluating the tree
using the posterior mean. This is illustrated empirically 
in the section below in the case of simple Bandit problems. 

%\clearpage
\paragraph{BAMCP versus Gittins indices}

\begin{figure}[hb]
\centering
\includegraphics[width=4.5in]{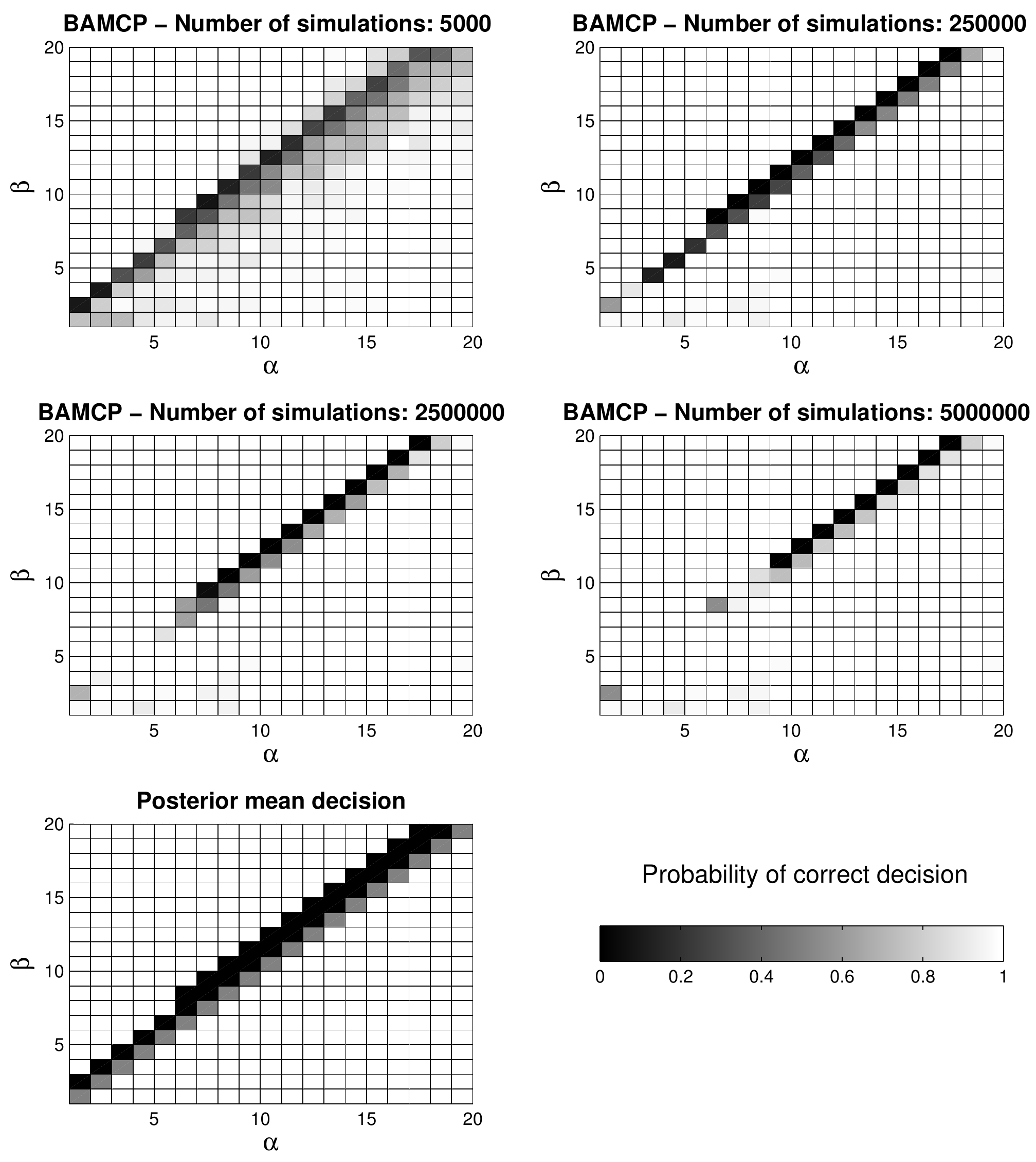}
\vspace{0.1in}
\caption{\small{Evaluation of BAMCP against the Bayes-optimal policy, for the
case $\gamma = 0.95$, when choosing between a deterministic arm with reward
$0.5$ and a stochastic arm with reward $1$ with posterior probability $p \sim
\text{Beta}(\alpha,\beta)$.
The result is tabulated for a range of values of $\alpha,\beta$, each cell
value corresponds to the probability of making the correct decision (computed
over 50 runs) when compared to the Gittins indices~\cite{Gittins:1989:aaa} for
the corresponding posterior. The first four tables corresponds to different
number of simulations for BAMCP and the last table shows the performance when
acting according to the posterior mean.
In this range of $\alpha,\beta$ values, the Gittins indices for the stochastic
arm are larger than $0.5$ (i.e., selecting the stochastic arm is optimal) for
$\beta \leq \alpha + 1$ but also $\beta = \alpha + 2$ for $\alpha \geq 6$.
Acting according to the posterior mean is different than the Bayes-optimal
decision when $\beta >= \alpha$ and the Gittins index is larger than $0.5$.
BAMCP is guaranteed to converges to the Bayes-optimal decision in all cases,
but convergence is slow for the edge cases where the Gittins index is close
to $0.5$ (e.g., For $\alpha=17, \beta=19$, the Gittins index is $0.5044$
which implies a value of $0.5044/(1-\gamma)=10.088$ for the stochastic arm
versus a value of $0.5 + \gamma \times 10.088 = 10.0836$ for the 
deterministic arm). 
}}
\label{fig:gittins095} 
\end{figure}

\begin{figure}[thp]
%\vspace{-0.15in}
\begin{center}
\subfigure[]{
\includegraphics[height=2.6in]{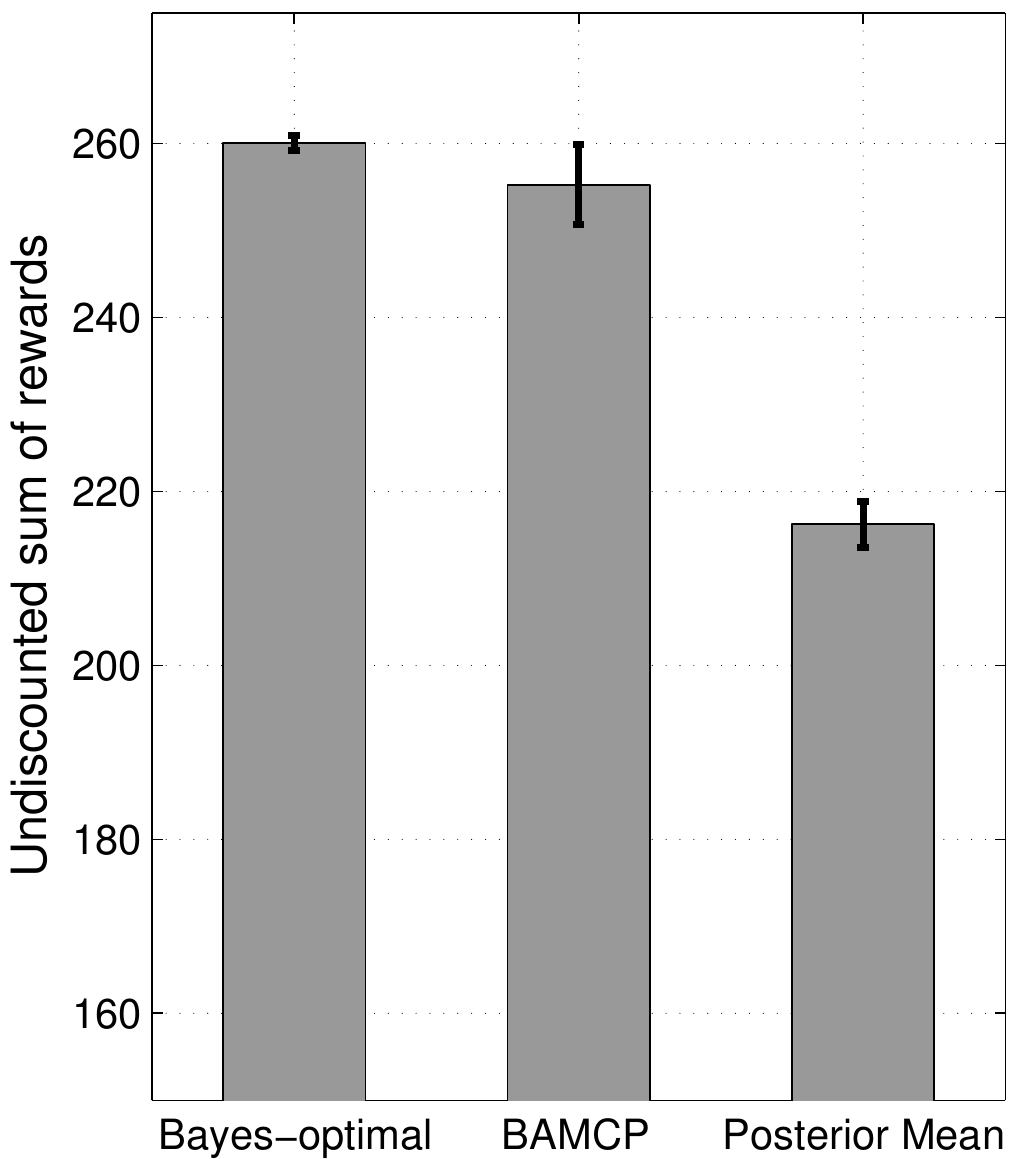}
\label{fig:gittins99_cumR}
}
\subfigure[]{
\includegraphics[height=2.6in]{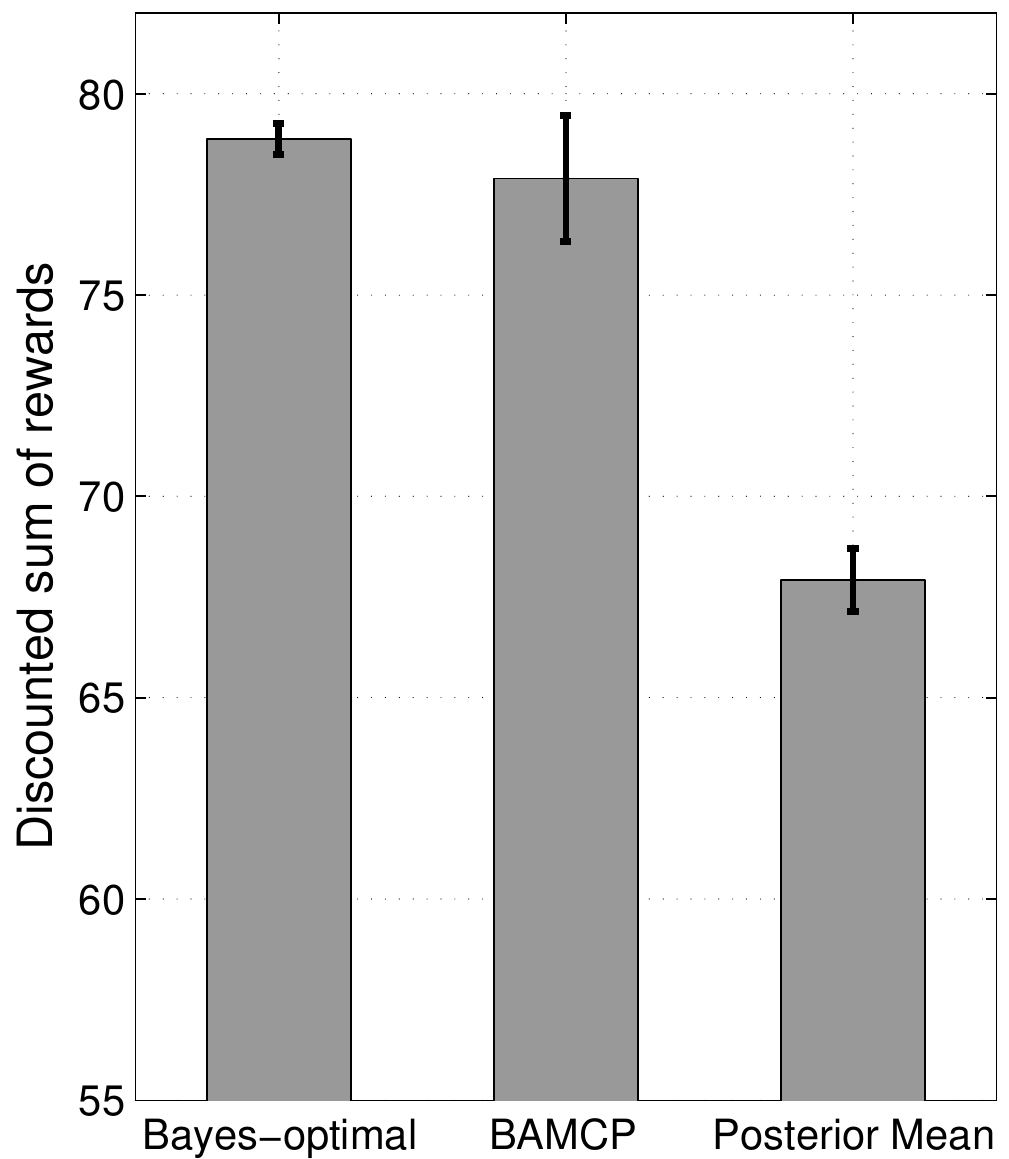}
\label{fig:gittins99_discR}
}
\vspace{-0.12in}
\caption{\small{Performance comparison of BAMCP (50000 simulations, 100 runs)
against the posterior mean decision on an 8-armed Bernoulli bandit with
$\gamma=0.99$ after 300 steps. The arms' success probability are
all $0.6$ except for one arm which has success probability $0.9$.
The Bayes-optimal result is obtained from 1000 runs with
the Gittins indices~\cite{Gittins:1989:aaa}. 
\textbf{a.} Mean sum of rewards after 300 steps.
\textbf{b.} Mean sum of discounted rewards after 300 steps.}} 
\label{fig:gittins99}
\end{center}
\vspace{-0.1in}
\end{figure}

\vspace{-0.2in}

\begin{figure}[hp!]
\begin{center}
\includegraphics[width=0.95\textwidth]{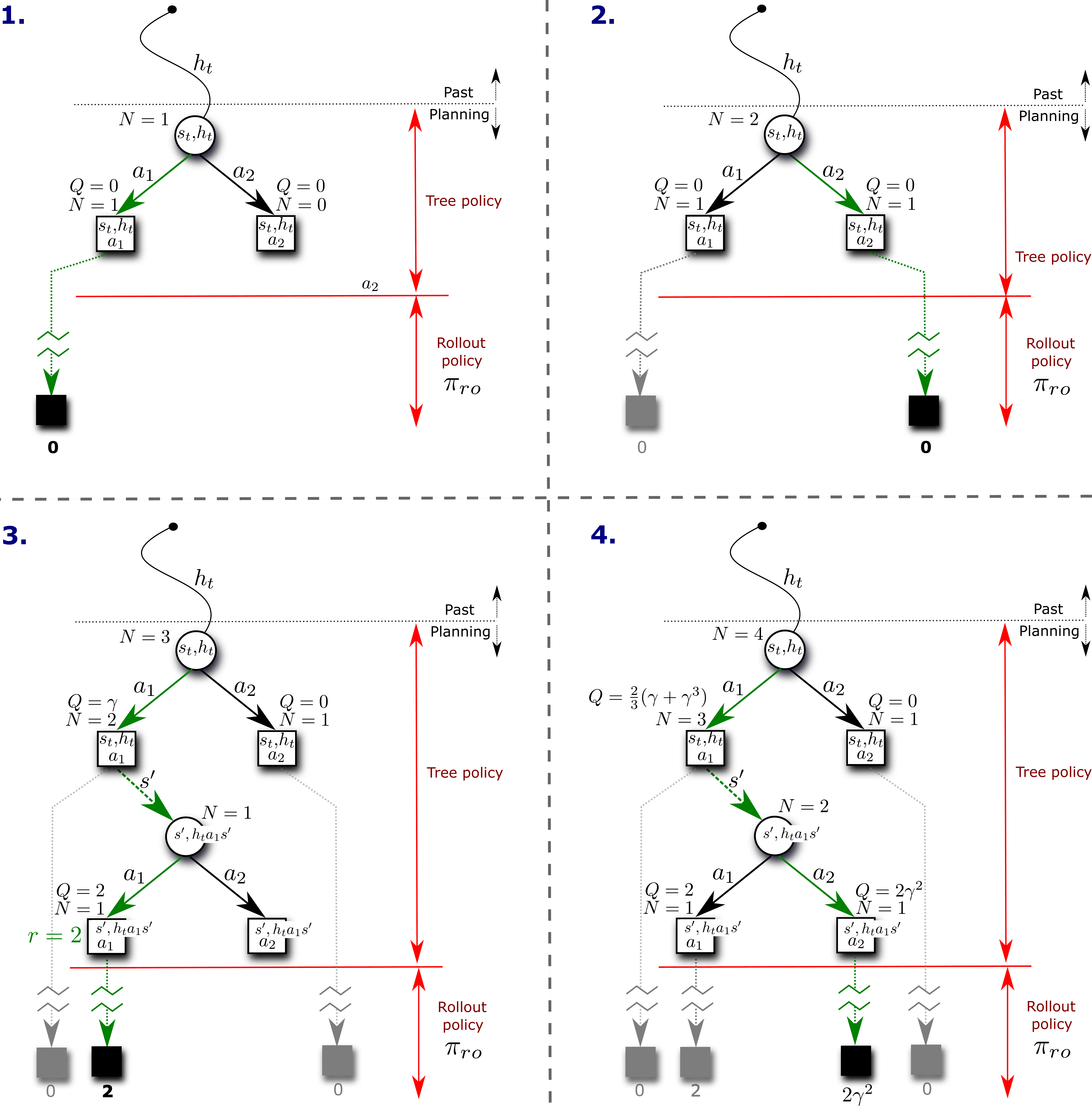}
\caption{\small{This diagram presents the first 4 simulations of BAMCP in an MDP
with 2 actions from state $\langle s_t, h_t \rangle$. The rollout trajectories
are represented with dotted lines (green for the current rollouts, and greyed
out for past rollouts). \textbf{1.} The root node is expanded with two action
nodes. Action $a_1$ is chosen at the root (random tie-breaking) and a rollout
is executed in $\mathcal{P}^1$ with a resulting value estimate of 0. Counts
$N(\langle s_t, h_t \rangle)$ and $N(\langle s_t, h_t \rangle, a_1)$, and value
$Q(\langle s_t, h_t \rangle, a_1)$ get updated. \textbf{2.} Action $a_2$ is
chosen at the root and a rollout is executed with value estimate 0. Counts and
value get updated. \textbf{3.} Action $a_1$ is chosen (tie-breaking), then $s'$
is sampled from $\mathcal{P}^3(s_t,a_1,\cdot)$. State node $\langle s', h_t a_1
s' \rangle$ gets expanded and action $a_1$ is selected, incurring a reward of
$2$, followed by a rollout. \textbf{4.} The UCB rule selects action $a_1$ at the top,
the successor state $s'$ is sampled from $\mathcal{P}^4(s_t,a_1,\cdot)$. Action
$a_2$ is chosen from the internal node $\langle s', h_ta_1s' \rangle$, followed
by a rollout using $\mathcal{P}^4$ and $\pi_{ro}$. A
reward of $2$ is obtained after $2$ steps from that tree node. Counts for the
traversed nodes are updated and the MC backup updates $Q(\langle s', h_ta_1s'
\rangle,a_1)$ to $R=0+\gamma 0+\gamma^2 2 + \gamma^3 0 + \dots =\gamma^2 2$ and
$Q(\langle s_t, h_t \rangle,a_1)$ to
$\gamma+\sfrac{2\gamma^3-\gamma}{3}=\frac{2}{3}(\gamma+\gamma^3)$. }} \label{fig:bmcpdi}
\end{center}
\end{figure}
%\vspace{-0.2in}

\begin{figure}[h!]
\centering

$\vdots$

$\cdots
\begin{array}{l}
\includegraphics[width=0.25\textwidth]{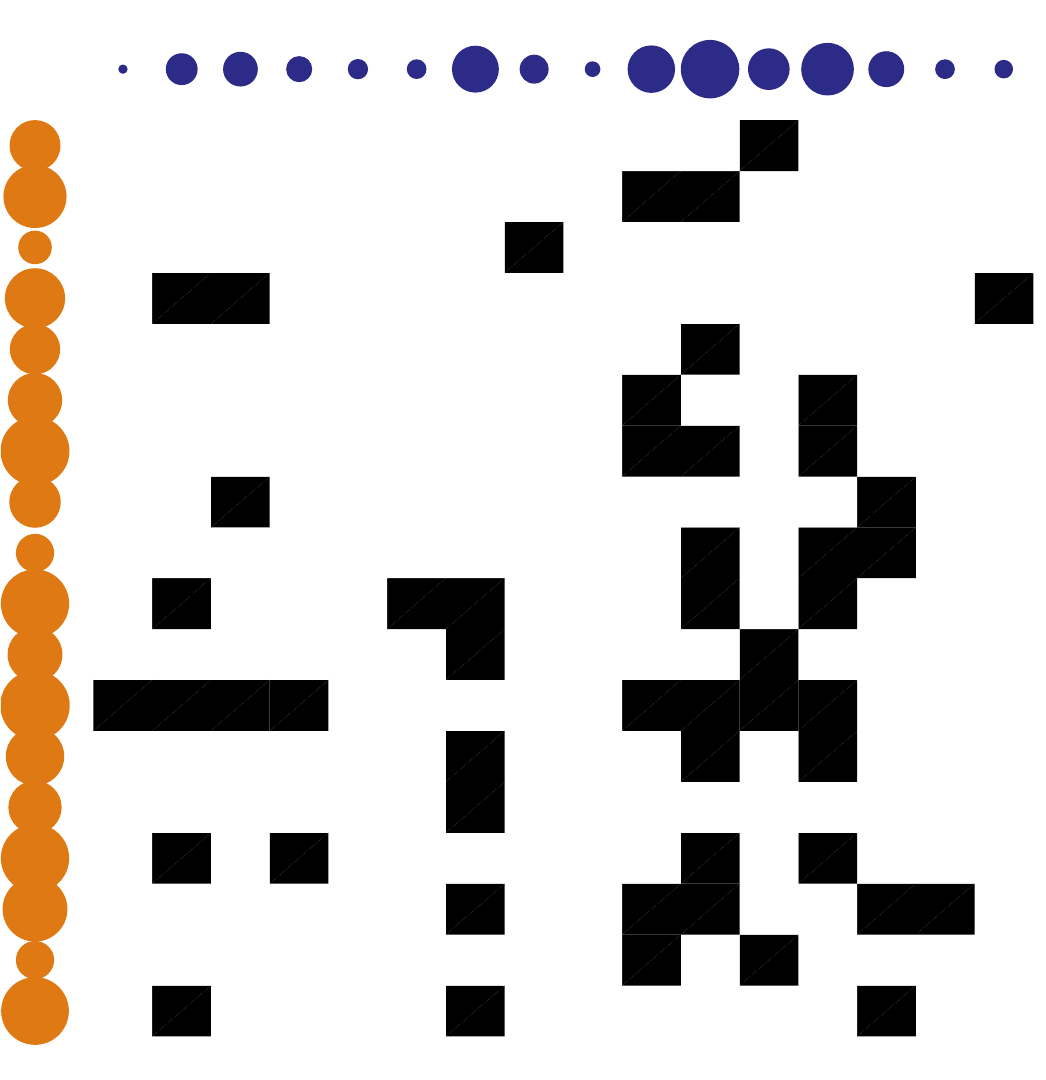}
\end{array}
\cdots$

$\vdots$
\vspace{-0.1in}
\caption{\small{A portion of an infinite 2D grid task generated with Beta
distribution parameters $\alpha_1=1,\beta_1=2$ (columns) and 
$\alpha_2=2,\beta_2=1$ (rows). Black squares at location (i,j) indicates
a reward of 1, the circles represent the corresponding parameters $p_i$ (blue)
and $q_j$ (orange) for each row and column (area of the circle is proportional
to the parameter value).
One way to interpret these parameters is that following column $i$ implies a
collection of $2 p_i / 3$ reward on average ($2/3$ is the mean of a Beta$(2,1)$
distribution) whereas following any row $j$ implies a collection of $q_j /
3$ reward on average; but high values of parameters $p_i$ are less likely than
high values parameters $q_j$.  These parameters are employed for the results
presented in Figure~\ref{fig:infgridset3}.}} \label{fig:infgridviz}
\end{figure}

\begin{figure}[h!]
\centering
\subfigure{
(a) \includegraphics[width=0.68\textwidth]{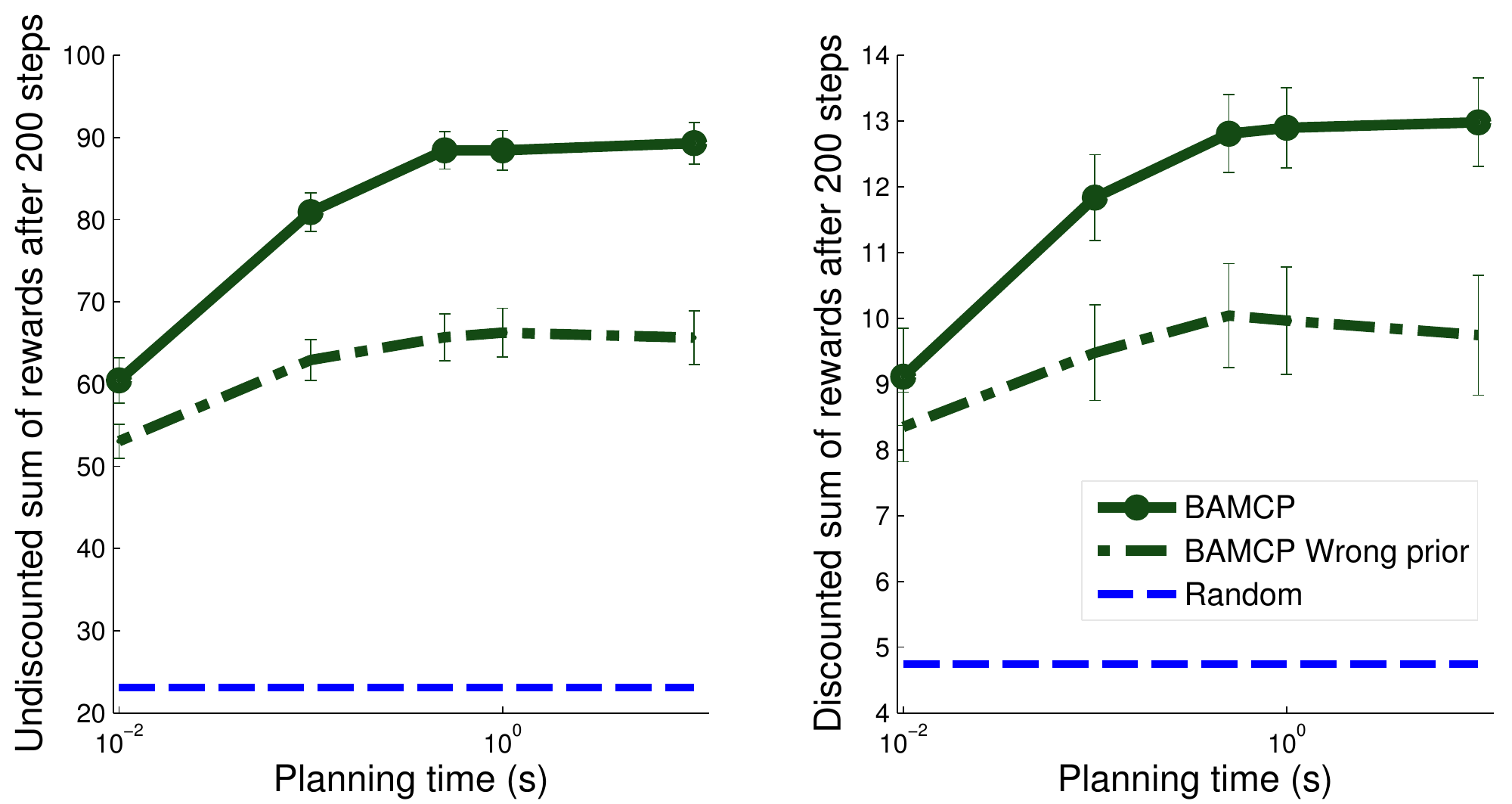}
}

(b) \subfigure{
\includegraphics[width=0.68\textwidth]{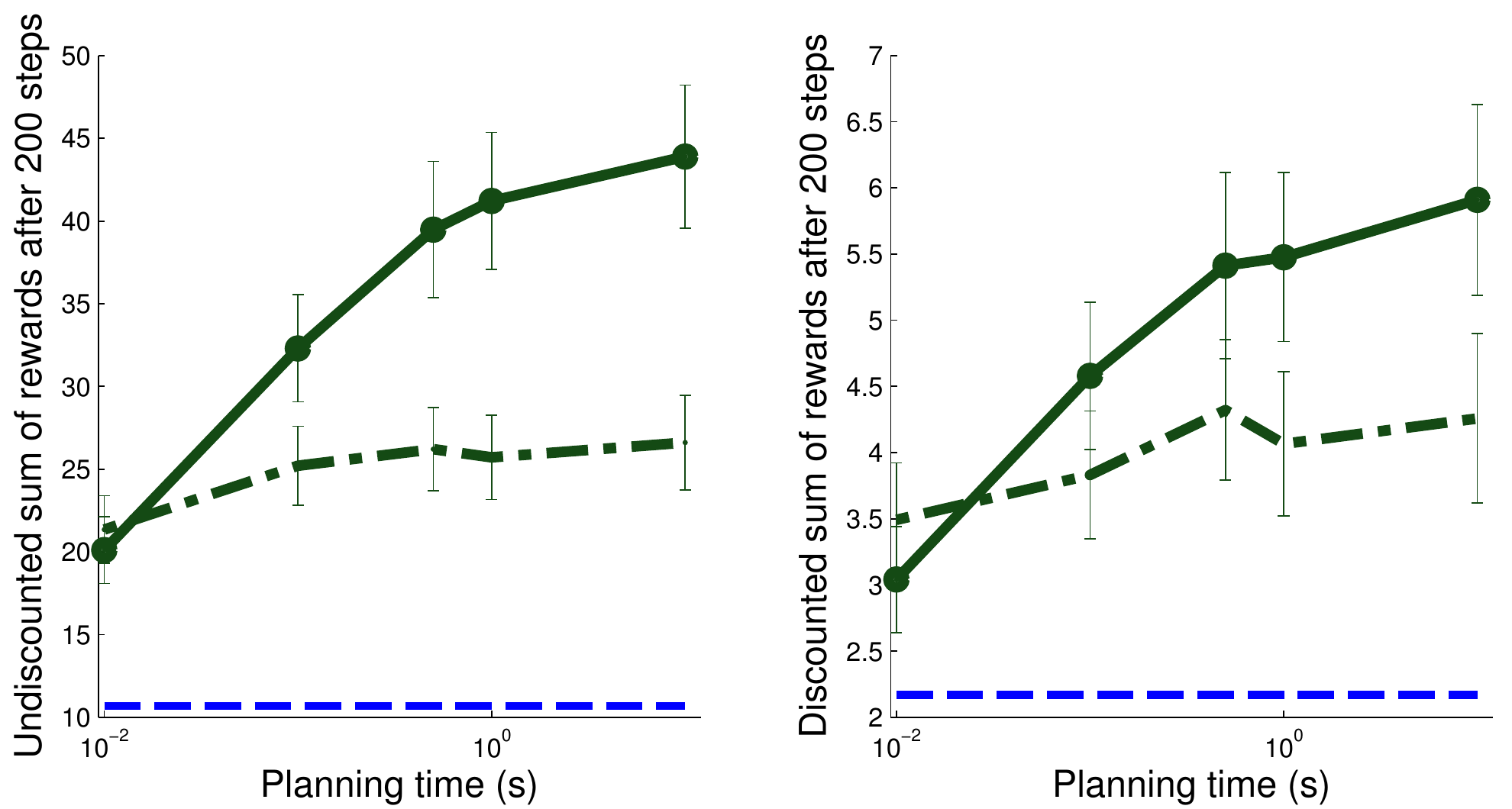}
}
\vspace{-0.1in}
\caption{\small{Performance of BAMCP on the Infinite 2D grid task of
Section~\ref{sec:infgrid}, for $\gamma=0.97$, as in
Figure~\ref{fig:infgridset3} but where the grids are generated with Beta
parameters \textbf{(a)} $\alpha_1=0.5,\beta_1=0.5,\alpha_2=0.5,\beta_2=0.5$ and
\textbf{(b)} $\alpha_1=0.5,\beta_1=0.5,\alpha_2=1,\beta_2=3$. In the wrong
prior scenario (green dotted line), BAMCP is given the parameters  \textbf{(a)}
$\alpha_1=4,\beta_1=1,\alpha_2=0.5,\beta_2=0.5$ and \textbf{(b)}
$\alpha_1=1,\beta_1=3,\alpha_2=0.5,\beta_2=0.5$. The behavior of the agent is
qualitatively different depending on the prior parameters employed (see
supplementary videos). For example, for the scenario in
Figure~\ref{fig:infgridset3}, rewards are often found in relatively dense blocks on
the map and the agents exploits this fact when exploring; for the scenario (b)
of this Figure, good rewards rates can be obtained by following the rare rows
that have high $q_j$ parameters, but finding good rows can be expensive so the
agent might settle on sub-optimal rows (as in Bandit problems where the
Bayes-optimal agent might settle on sub-optimal arm if it believes it likely is
the best arm given past data). It should be pointed out that the actual
Bayes-optimal strategy in this domain is not known --- the behavior of BAMCP for
finite planning time might not qualitatively match the Bayes-optimal
strategy.}} 
\label{fig:infgridset12}
\vspace{-0.2in}
\end{figure}

\paragraph{Inference details for the infinite 2D grid task of Section~\ref{sec:infgrid}}
\vspace{-0.1in}
We construct a Markov Chain using the Metropolis-Hastings algorithm to sample
from the posterior distribution of row and column parameters given observed
transitions, following the notation introduced in Section~\ref{sec:infgrid}.
Let $O = \{ (i,j) \}$ be the set of observed reward locations, each associated
with an observed reward $r_{ij} \in \{0,1\}$.  The proposal distribution
chooses a row-column pair $(i_p,j_p)$ from $O$ uniformly at random, and samples
$\tp_{i_p} \sim \text{Beta}(\alpha_1+ m_1, \beta_1 + n_1)$ and $\tq_{j_p} \sim
\text{Beta}(\alpha_2+ m_2, \beta_2 + n_2)$, where $m_1 = \sum_{(i,j) \in O}
\mathbf{1}_{i = i_p} r_{ij}$ (i.e., the sum of rewards observed on that column)
and $n_1 = (1 - \sfrac{\beta_2}{2(\alpha_2+\beta_2)}) \sum_{(i,j) \in
O}\mathbf{1}_{i = i_p} (1-r_{ij})$, and similarly for $m_2,n_2$ (mutatis
mutandis).  The $n_1$ term for the proposed column parameter $\tp_{i}$ has this
rough correction term, based on the prior mean failure of the row parameters,
to account for observed $0$ rewards on the column due to potentially low row
parameters. Since the proposal is biased with respect to the true conditional
distribution (from which we cannot sample), we also prevent the proposal
distribution from getting too peaked.  Better proposals (e.g., taking into
account the sampled row parameters) could be devised, but they would likely
introduce additional computational cost and the proposal above generated large
enough acceptance probabilities (generally above $0.5$ for our experiments).
All other parameters $p_i,q_j$ such that $i$ or $j$ is present in $O$ are kept from
the last accepted samples (i.e., $\tp_i = p_i$ and $\tq_j = p_j$ for these $i$s
and $j$s), and all parameters $p_i,q_j$ that are not linked to observations are
(lazily) resampled from the prior --- they do not influence the acceptance
probability. We denote by $Q(\bp,\bq \rightarrow \tbp,\tbq)$ the probability of
proposing the set of parameters $\tbp$ and  $\tbq$ from the last accepted
sample of column/row parameters $\bp$ and $\bq$. 
The acceptance probability $A$ can then be computed as $A = \min (1, A')$
where:
\begin{align*}
	A' &= \frac{P( \tbp,\tbq | h) Q(\tbp,\tbq \rightarrow \bp,\bq)}{ P( \bp,\bq | h) Q(\bp,\bq \rightarrow \tbp,\tbq)} \\
	   &= \frac{P(\tbp,\tbq) Q(\tbp,\tbq \rightarrow \bp,\bq)P(h | \tbp,\tbq)}{
		          P(\bp,\bq) Q(\bp,\bq \rightarrow \tbp,\tbq) P(h | \bp,\bq)} \\
	 &=	\frac{p_{i_p}^{m_1} (1-p_{i_p})^{n_1} 
	 q_{j_p}^{m_2} (1-q_{j_p})^{n_2}
	 \prod_{(i,j) \in O} \mathds{1}[i = i_p \text{ or } j = j_p] (\tp_i \tq_j)^{r_{ij}} (1- \tp_i \tq_j)^{1-r_{ij}}}{
	 \tp_{i_p}^{m_1} (1-\tp_{i_p})^{n_1} 
	 \tq_{j_p}^{m_2} (1-\tq_{j_p})^{n_2}
	 \prod_{(i,j) \in O} \mathds{1}[i = i_p \text{ or } j = j_p] (p_i q_j)^{r_{ij}} (1- p_i q_j)^{1-r_{ij}}}.
\end{align*}
The last accepted sampled is employed whenever a sample is rejected. Finally,
reward values $R_{ij}$ are resampled lazily based on the last accepted sample of
the parameters $p_i,q_j$, when they have not been observed already.  We
omit the implicit deterministic mapping to obtain the dynamics $\mP$ from these
parameters. 

\end{document}